\documentclass{article}

\PassOptionsToPackage{numbers, compress}{natbib}

\usepackage[preprint]{neurips_2025}

\usepackage[utf8]{inputenc} 
\usepackage[T1]{fontenc}    
\usepackage{hyperref}       
\usepackage{url}            
\usepackage{booktabs}       
\usepackage{amsmath}        
\usepackage{amsfonts}       
\usepackage{nicefrac}       
\usepackage{microtype}      
\usepackage{xcolor}         
\usepackage{graphicx}
\usepackage{subcaption} 
\usepackage{multirow}
\usepackage{array}
\title{Abstract-LoRA: Unlocking Single-Image Style Transfer through Targeted U-Net Block Training
}

\author{
\textbf{Xinglin Hu} \\
School of Data Science, The Chinese University of Hong Kong, Shenzhen \\
{\tt\small 123090187@link.cuhk.edu.cn}
}

\begin{document}

\maketitle

\begin{abstract}
Diffusion models represent one of the most advanced paradigms in generative modeling. Leveraging their development, a growing number of style transfer methods based on diffusion models have been proposed. However, among these methods, multi-image style transfer approaches that require at least five to ten style examples tend to achieve more satisfactory results. Single-image methods, by contrast, often struggle with either insufficient content preservation or inadequate style fidelity. This greatly limits style extraction from scarce artworks and undermines their artistic value. To address this, we propose Abstract-LoRA, a method that pushes the boundaries of single-image style transfer through lightweight LoRA training on specific U-Net blocks in diffusion models. Specifically, our work is inspired by B-LoRA, a style transfer method that achieves basic style-content disentanglement by training specific U-Net blocks. However, it suffers from a critical limitation—the inability to capture complex backgrounds. Building upon B-LoRA, our method conducts a more refined analysis of U-Net blocks, employing additional U-Net blocks and clustering-based abstraction of style images to better disentangle and balance style and content. Extensive experiments demonstrate that our proposed method not only generates visually more harmonious and satisfying artistic images but also quantitatively improves the preservation of both style and content in the final outputs.
\end{abstract}

\section{Introduction}
Artistic style transfer has been a long-standing research topic. Since Gatys et al. \cite{gatys2016} first proposed a method based on pre-trained deep convolutional neural networks (CNNs), style transfer has continued to develop rapidly, largely driven by the evolution of fundamental paradigms—from CNNs \cite{lecun1989} to VAEs \cite{kingma2014}, GANs \cite{goodfellow2014}, diffusion models \cite{ho2020}, and more recently, flow matching \cite{lipman2023} and autoregressive models \cite{vandenoord2016}, each accompanied by corresponding style transfer methods \cite{gatys2016,upchurch2016,karras2019,ruiz2023,an2021,nguyen2025}. Among these paradigms, diffusion models offer the best overall combination of advancement and maturity, and our research builds upon diffusion models to further exploit their potential in style transfer.

Numerous mature diffusion-based style transfer methods have already emerged \cite{ruiz2023,li2024,chung2024,zhang2023}, which complement and integrate with each other, progressively enhancing the quality of style transfer results. Methods that draw on five to ten style examples per target style tend to perform best, since multiple exemplars let the model average out content-specific details and converge on statistics that are genuinely characteristic of the style. With only a single style image, this averaging is not possible: models risk entangling the style image's own incidental content with its style, which in practice manifests as either leaked content from the style image (compromising content preservation) or an overly conservative fit that under-represents the target style (compromising style fidelity). This trade-off substantially limits style extraction from scarce artworks that exist as unique, non-reproducible pieces. Considering this limitation, we aim to push the boundaries of single-image style transfer with diffusion models, specifically employing SDXL \cite{podell2023}.

Our work builds directly on B-LoRA \cite{frenkel2024}, which achieves basic style-content disentanglement by assigning distinct U-Net blocks to style and content during LoRA training. This block-wise separation is effective for foreground-dominant images, but it treats all non-style content as a single undifferentiated block, leaving it unable to capture complex backgrounds—an imbalance we identify as the central obstacle to extending B-LoRA to more diverse, real-world compositions.

To address this issue, we decompose content into two complementary components: background and subject. Through a fine-grained analysis of the U-Net architecture in SDXL, we identify specific U-Net blocks that are strongly associated with these two components. For style modeling, we follow the same U-Net blocks adopted in B-LoRA. By injecting content image information into two distinct LoRA modules simultaneously, as illustrated in Figure~\ref{fig:arch-evolution}(b), our method effectively preserves both background structures and subject details during stylization.

However, this approach introduces another problem: due to the complex background structure and strong background information, certain features from the style image are weakened—specifically, its color characteristics (hue, brightness, saturation), color proportions, and color separation (i.e., the distinctness between different color regions, which tends to blur as colors bleed across the newly introduced background structure). Naively scaling up the style image's LoRA weights does not resolve this issue, as we analyze in detail in Section~3.3. 

To address this, we propose a corresponding strategy: we extract an abstract representation from the style image, composed of adjoining square color blocks whose sizes reflect each color's relative dominance. This abstract representation purely contains the weakened features from the style image. By injecting this representation into the style-responsible U-Net block in the same manner, as illustrated in Figure~\ref{fig:arch-evolution}(c), we restore these features with almost no cost of content.

Through experiments on various types of style and content images, we find that our method generally improves upon existing single-image style transfer approaches such as B-LoRA in visual quality, achieving a more favorable overall balance between style and content preservation.

\section{Related Work}

\noindent\textbf{Style transfer}\quad Image style transfer has been a perennial challenge in computer vision, aiming to alter the style of an input image based on a given reference. With the evolution of fundamental paradigms—from CNNs to VAEs, GANs, diffusion models, and more recently flow matching and autoregressive models—numerous studies have explored their potential and developed corresponding style transfer methods. 

Gatys et al. use a pre-trained VGG network to extract content features (high-level features) and style features (Gram matrices of multi-layer features), then optimize a random noise image to minimize both content loss and style loss, achieving style transfer \cite{gatys2016}. ST-VAE projects nonlinear styles into a linear latent space where multiple styles can be merged via interpolation before transferring to content images \cite{liu2021}. GANILLA uses concatenative skip connections in downsampling and long skip connections in upsampling to merge low-level and high-level features, achieving better style-content balance in a CycleGAN-based framework \cite{hicsonmez2020}. DreamBooth can be applied for style transfer by fine-tuning a diffusion model on a few reference images of a specific artistic style (instead of a subject), binding the style to a unique text identifier (e.g., ``[V] style''), then generating new images in that learned style through text prompts \cite{ruiz2023}. ArtFlow uses reversible neural flows (GLOW) to losslessly project content and style images into latent feature space, performs style transfer in feature space (via AdaIN or WCT), then reconstructs stylized images through backward propagation \cite{an2021}. CSD-VAR performs style transfer by using Visual Autoregressive Modeling's scale-wise generation process to disentangle content and style representations through alternating optimization at different scales, with SVD-based rectification preventing content leakage and augmented key-value memory preserving content identity during stylization \cite{nguyen2025}. 

Among existing generative modeling paradigms, diffusion models stand out as the most advanced and mature framework in terms of both technical sophistication and practical deployment, consequently giving rise to a relatively richer ecosystem of high-quality style transfer methods. Nevertheless, single-image style transfer approaches based on diffusion models frequently suffer from either insufficient content preservation or inadequate style fidelity. This study addresses these challenges through systematic decomposition and strategic exploitation of the U-Net architecture, thereby pushing the boundaries of single-image style transfer with diffusion models.

\noindent\textbf{LoRA for Image Stylization}\quad LoRA is commonly employed for image stylization by fine-tuning models to generate images with specific styles. Typically, a LoRA is trained on a set of images and then combined with control methods such as Concept-Sliders \cite{gandikota2023} or ControlNet \cite{zhang2023controlnet}, along with text prompts to condition the content of generated images. Conventionally, capturing style and capturing content each require a separately trained LoRA, with the two optimized toward distinct objectives yet covering the same full set of network parameters; naively combining them therefore causes interference, with no straightforward method to merge them cleanly.

Recently, however, Shah et al. proposed ZipLoRA \cite{shah2023}, which reconciles this conflict by learning per-column mixing coefficients that merge two independently trained style and content LoRAs into a unified ``zipped'' LoRA. This novel paradigm substantially enhances the potential and efficiency of LoRA-based methods in style transfer. B-LoRA \cite{frenkel2024} instead sidesteps the merging problem altogether: by training both LoRAs with the same generic reconstruction objective, it finds that individual U-Net blocks naturally specialize toward style or content, allowing the relevant block from each LoRA to simply be assembled together without any learned merging step, thereby achieving basic style-content disentanglement. Our method extends and refines this block-wise paradigm through more detailed U-Net block analysis and abstract representation extraction, enabling LoRA training to capture and preserve richer, more nuanced information, thereby achieving better balance and integration between style and content.

\section{Method}

\begin{figure}[t]
    \centering
    \includegraphics[width=1.0\linewidth]{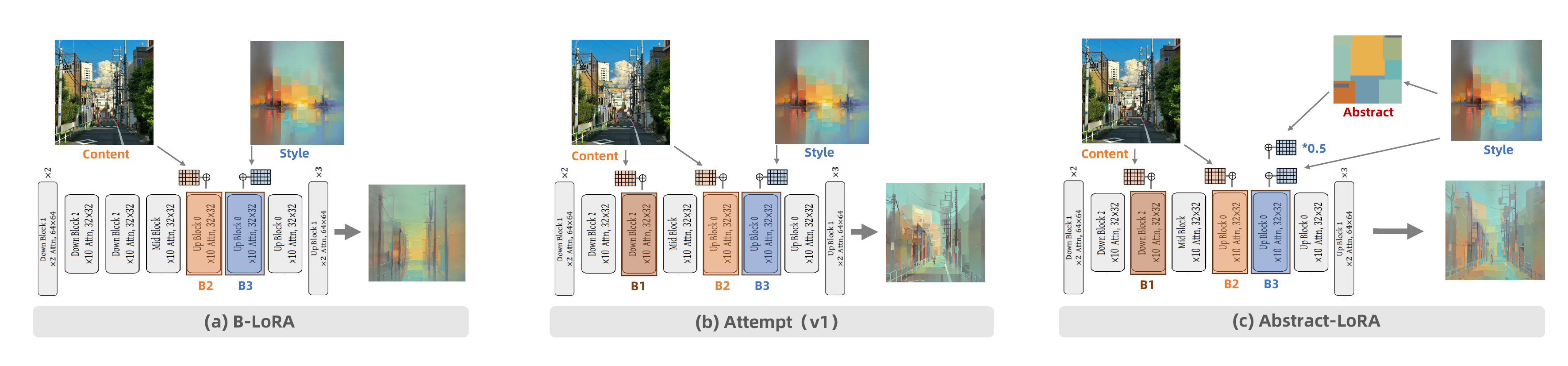}
    \caption{Architecture evolution from B-LoRA to Abstract-LoRA. (a) B-LoRA injects LoRA weights only at B2 and B3, leaving background content unconstrained. (b) Attempt (v1) adds an additional injection at B1 to capture background information, improving content preservation at the cost of style fidelity. (c) The final Abstract-LoRA further injects an abstract color representation (scaled by 0.5) into B3, recovering style fidelity with minimal cost to content.}
    \label{fig:arch-evolution}
\end{figure}

We first disentangle the style and content of input images into independent components (distinct LoRA weights) through LoRA training. Then, during the generation process, we integrate the content components from the content image with the style components from the style image to achieve style-content fusion. The key aspect of this paradigm lies in where to inject the LoRA weights. B-LoRA \cite{frenkel2024} injects LoRA weights at only two locations (B2 and B3), which is insufficient for effectively capturing the background information from the content image, as demonstrated in Figure~\ref{fig:arch-evolution}(a). In contrast, our method introduces an additional LoRA injection point, further disentangling the subject and background of the content image to achieve more complete and comprehensive content preservation. Moreover, we extract an abstract representation from the style image to recover the style features that are weakened due to the introduction of additional content LoRA weights. This design philosophy motivates the name of our method: ``Abstract-LoRA''.

\subsection{SDXL Architecture Analysis}

Similar to previous works \cite{voynov2023,agarwal2025}, we investigate how different layers within the base text-to-image model affect the generated output. Following a strategy similar to Voynov et al. \cite{voynov2023}, the core idea is to inject an alternative text prompt $\hat{p}$ into the cross-attention layers of a specific Transformer block in SDXL, while keeping the original prompt $p$ for all remaining blocks, and then examine whether this localized modification produces a noticeable change in the final image. A significant change indicates that the corresponding block is strongly associated with the attribute altered between $p$ and $\hat{p}$.

In B-LoRA \cite{frenkel2024}, experiments have shown that block B2 is highly correlated with content, whereas block B3 is more related to style. Building on this observation, we further investigate whether the content itself can be decomposed into two components---foreground subjects and background.

To this end, we construct multiple prompt pairs to identify which blocks are responsible for background information and which are responsible for the main subject. By visually inspecting the generated outputs and further quantifying the effect through cosine similarity between CLIP embeddings of the result images and prompts \cite{radford2021}, we obtain the following conclusion: B1 is strongly correlated with background, while B2 is strongly correlated with the subject.

\begin{figure}[t]
    \centering
    \includegraphics[width=1\linewidth]{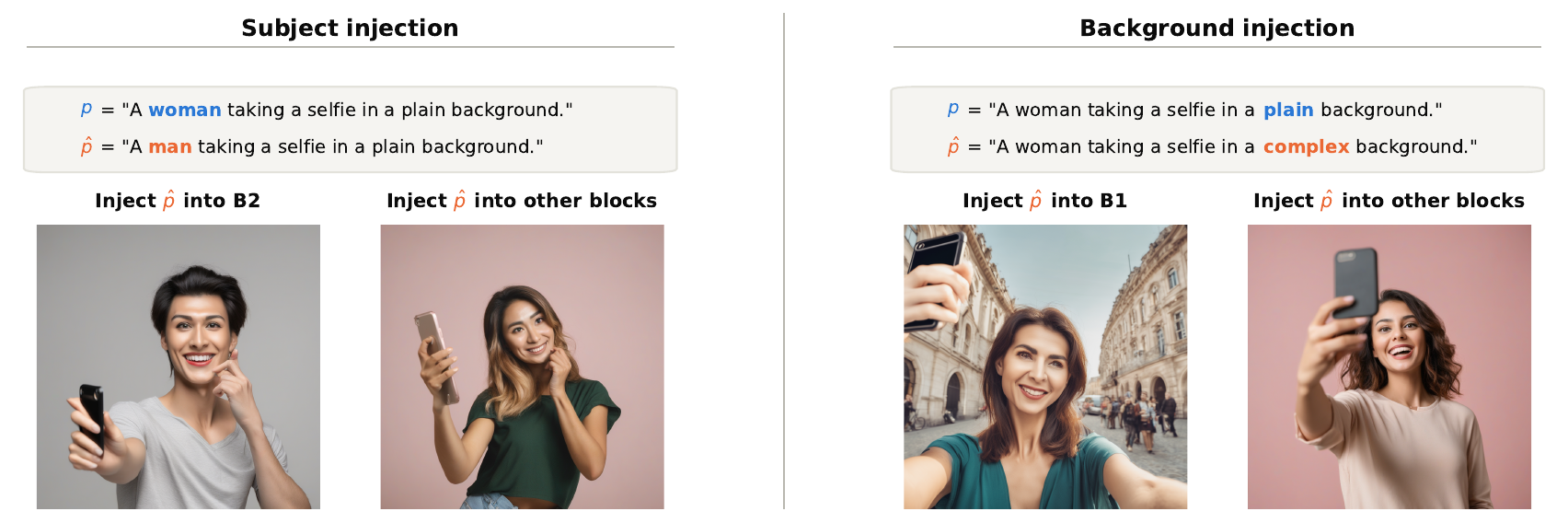}
    \caption{Prompt injection effect on the generated image. In the subject-injection test (left), injecting $\hat{p}$ into B2 changes the subject while the background remains unaffected. In the background-injection test (right), injecting $\hat{p}$ into B1 changes the background while the subject remains unaffected.}
    \label{fig:block-analysis}
\end{figure}

Figure~\ref{fig:block-analysis} illustrates representative examples. In the subject-injection test (left), we use the prompt pair 
$p =$ ``A woman taking a selfie in a plain background'' and 
$\hat{p} =$ ``A man taking a selfie in a plain background'' 
to identify subject-related blocks. In the background-injection test (right), we use 
$p =$ ``A woman taking a selfie in a plain background'' and 
$\hat{p} =$ ``A woman taking a selfie in a complex background'' 
to identify background-related blocks. The resulting images are consistent with our findings.

\subsection{Disentanglement and Fusion}

\begin{figure}[t]
    \centering
    \includegraphics[width=1.0\linewidth]{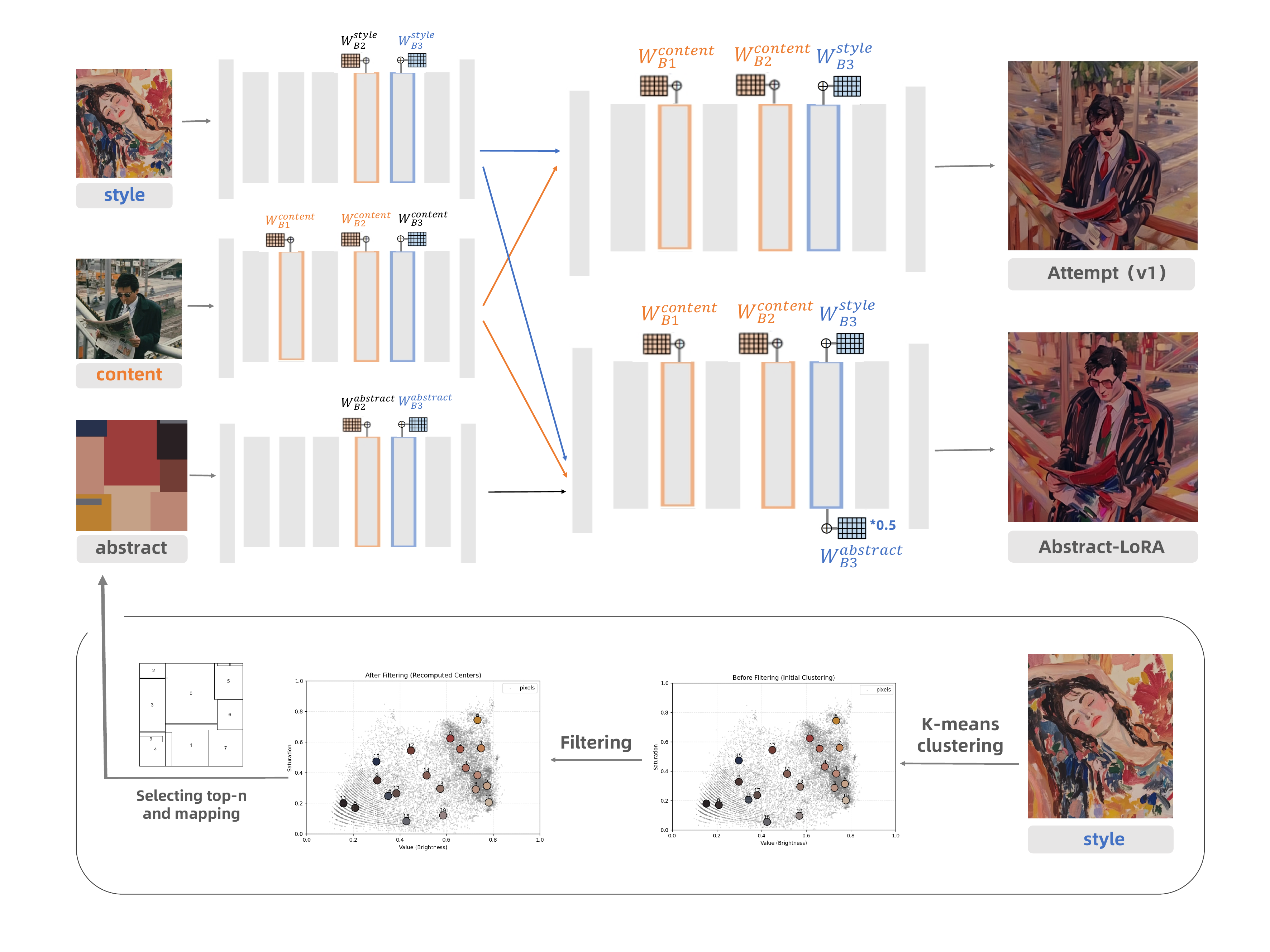}
    \caption{Detailed content--style--abstract disentanglement mechanism. Top: the style, content, and abstract-representation images are each decomposed into per-block LoRA weights; the content-derived weights ($W_{B1}^{content}$, $W_{B2}^{content}$) and the style-derived weight ($W_{B3}^{style}$) are combined to form Attempt (v1), and further combined with the abstract representation's weight ($W_{B3}^{abstract}$, scaled by 0.5) to form the final Abstract-LoRA. Bottom: pipeline for extracting the abstract representation from the style image via K-means clustering, filtering, and top-$n$ color selection and mapping.}
    \label{fig:detailed-mechanism}
\end{figure}

Building on our analysis of the U-Net blocks, we further perform LoRA training to disentangle the style and content information from the input images. During inference, the style LoRA weights derived from the style image and the content LoRA weights derived from the content image are added to the attention weights of the corresponding U-Net blocks, allowing for fusion of style and content. As shown in Figure~\ref{fig:detailed-mechanism}, we disentangle information of the style image into LoRA weights $W_{B2}^{style}$ and $W_{B3}^{style}$, where $W_{B2}^{style}$ carries the content information present within the style image and $W_{B3}^{style}$ carries style information. For the content image, we inject information into the LoRA weights $W_{B1}^{content}$, $W_{B2}^{content}$ and $W_{B3}^{content}$, where $W_{B1}^{content}$ carries background information, $W_{B2}^{content}$ carries subject information, and $W_{B3}^{content}$ carries the style information inherent to the content image. 

After training, only the LoRA weights $W_{B1}^{content}$, $W_{B2}^{content}$, and $W_{B3}^{style}$ are applied to the corresponding U-Net blocks during generation; $W_{B2}^{style}$ and $W_{B3}^{content}$ are excluded, as incorporating them would reintroduce the style image's own subject content or the content image's own inherent style, respectively. This selective combination enables the generated images to integrate content and style information from the respective input images. Our method achieves substantially better background preservation compared to B-LoRA \cite{frenkel2024}. However, the strength of the transferred style is noticeably weakened. This limitation motivates our second improvement, which leverages abstract representations to recover style-specific characteristics of the style image. We refer to this intermediate version—prior to incorporating the abstract representation—as Attempt (v1) in our experimental comparisons (Section~4).

\subsection{Further Balancing via Abstract Representations}

\begin{figure}[t]
    \centering
    \includegraphics[width=1\linewidth]{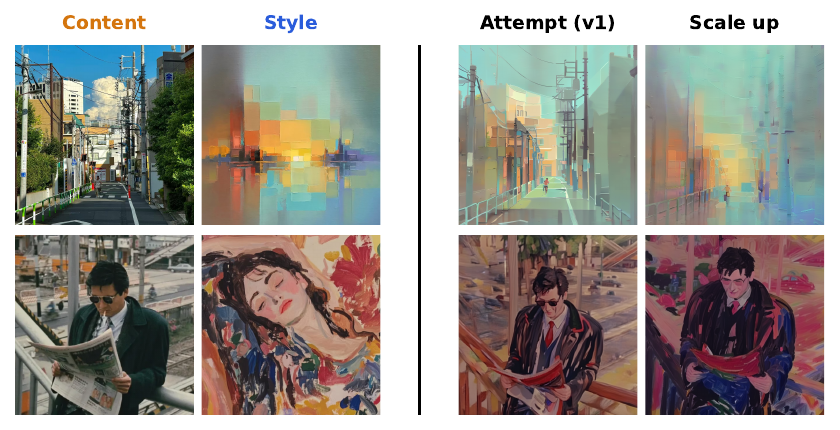}
    \caption{Comparison between Attempt (v1) and naively scaling up the style image's LoRA weights by a factor of 2, across two representative content--style pairs. Increasing the weight intensifies certain stylistic characteristics but also introduces noticeable content distortion and disrupts style-content disentanglement, confirming that this issue cannot be resolved through weight scaling alone.}
    \label{fig:scaleup}
\end{figure}

Upon analysis, we find that the weakened properties of the style image are mainly its color characteristics (hue, brightness, saturation), color proportions, and color separation. Moreover, this issue cannot be resolved simply by scaling up the style image's LoRA weights, as doing so would only intensify the conflict between style and content, resulting in worse style-content disentanglement and preservation, as shown in Figure~\ref{fig:scaleup}.

To address this, our method first extracts an abstract representation from the original style image, capturing only its color characteristics, color proportions, and color separation. Then, using the same approach, we apply LoRA training on the resulting abstract image, and subsequently integrate the learned weights into the corresponding U-Net layers during generation. This enables the style to be recovered with minimal loss of content.

\subsubsection{Extracting Abstract Representation}

The overall pipeline for extracting the abstract representation is illustrated in the lower portion of Figure~\ref{fig:detailed-mechanism}. Let an input style image be $I \in \mathbb{R}^{H \times W \times 3}$. The image is first downsampled to a smaller resolution $\tilde{I}$ for efficiency, and converted to the HSV color space:
\[
\tilde{I}_{\text{HSV}} = \text{RGB2HSV}(\tilde{I}) \in [0,1]^{\tilde{H} \times \tilde{W} \times 3}.
\]

Then, we cluster the pixels using the K-Means method \cite{lloyd1982} in the original RGB space:
\[
\{C_i\}_{i=1}^{k} = \text{KMeans}(\tilde{I}, k),
\]

Since the original cluster centroids may be biased by dim pixels, we recompute each centroid using only the pixels within the cluster that meet minimum brightness and saturation thresholds, making the dominant colors more vivid and representative. However, if fewer than 50\%  of a cluster's pixels meet these thresholds (including the case where no pixels meet them), we use all pixels in that cluster to ensure stable centroid computation and preserve the original clustering structure for inherently desaturated or dim color regions:

\[
C_i^\text{refined} = \frac{1}{|P_i^\text{high}|} \sum_{p \in P_i^\text{high}} p,
\]
where
\[
P_i^\text{high} = \{ p \in P_i \mid S(p) > s_\text{min} \land V(p) > v_\text{min} \}.
\]

After this, each cluster is assigned a weighted score combining its pixel count, average saturation, and brightness:
\[
\text{score}(C_i) = |P_i| \cdot \Big(1 + \alpha \cdot \bar{S}_i + \beta \cdot \bar{V}_i \Big),
\]
where $\bar{S}_i, \bar{V}_i$ are the mean saturation and brightness of cluster $i$, and $\alpha, \beta$ are weighting coefficients. Clusters are sorted by score, and the top-$n$ clusters are selected as dominant colors.

To ensure perceptual diversity, we enforce a minimum hue distance $\Delta H_\text{min}$ between selected clusters. Denoting the hue of cluster $i$ as $H_i$, we select clusters such that:
\[
\forall i,j \in \text{selected}, \quad \min(|H_i - H_j|, 360 - |H_i - H_j|) > \Delta H_\text{min}.
\]

If not enough clusters satisfy this criterion, $\Delta H_\text{min}$ is iteratively reduced.

Finally, the top-$n$ colors ${C_i^\text{top}}_{i=1}^{n}$ are mapped to predefined rectangular regions ${R_j}$ to produce the abstract color representation. The proportions and ordering of these rectangular regions are designed following empirical findings from art-historical analysis, which show that effective artworks typically exhibit a structured hierarchy of dominant and subordinate colors \cite{lee2018}. This design choice ensures that our abstract representation adheres to established aesthetic principles and preserves meaningful stylistic structure.

The resulting image retains only the dominant colors and their proportions, with hard boundaries between them.

\subsubsection{Inject Abstract Representation into Style Block}

After obtaining the abstract representation, we disentangle and inject its information in the same manner as for the style image. The only difference is that, before applying it to the U-Net weights, we scale its LoRA weights by 0.5, ensuring that it provides meaningful guidance while keeping potential side effects minimal. At this stage, the procedure aligns with the mechanism illustrated in Figure~\ref{fig:detailed-mechanism}.

Through this additional balancing provided by the abstract representation, we recover the target style with almost no loss of content preservation, as demonstrated in Figure~\ref{fig:arch-evolution}(c) and Figure~\ref{fig:detailed-mechanism}.

\section{Experimental Results}

In this section, we first introduce the experimental settings. Then we present quantitative comparisons between the proposed method and several baseline models.

\subsection{Experimental Settings}

\noindent\textbf{Implementation details}\quad We train LoRA weights (rank r = 64) on SDXL v1.0 \cite{podell2023} while keeping the model weights and text encoders frozen. Each single-image training uses Adam optimizer (lr = 5e-5), B-LoRA's default prompt, 1000 steps, and takes approximately 2 minutes on a single RTX 5090 GPU.

\noindent\textbf{Datasets}\quad Following \cite{huang2017,zhang2019,jing2020,chen2021}, we use MS-COCO \cite{lin2014} and WikiArt \cite{karayev2013} as content and style datasets, respectively. During training, we resize all images to 512×512 to ensure resolution consistency.

\noindent\textbf{Complex Background Scenarios}\quad We evaluate our method across four challenging scenarios: (1) architectural complex backgrounds, (2) natural complex backgrounds, (3) single-subject complex backgrounds, and (4) multi-object complex backgrounds.

\subsection{Quantitative Comparisons}

\begin{figure}[t]
    \centering
    \includegraphics[width=1\linewidth]{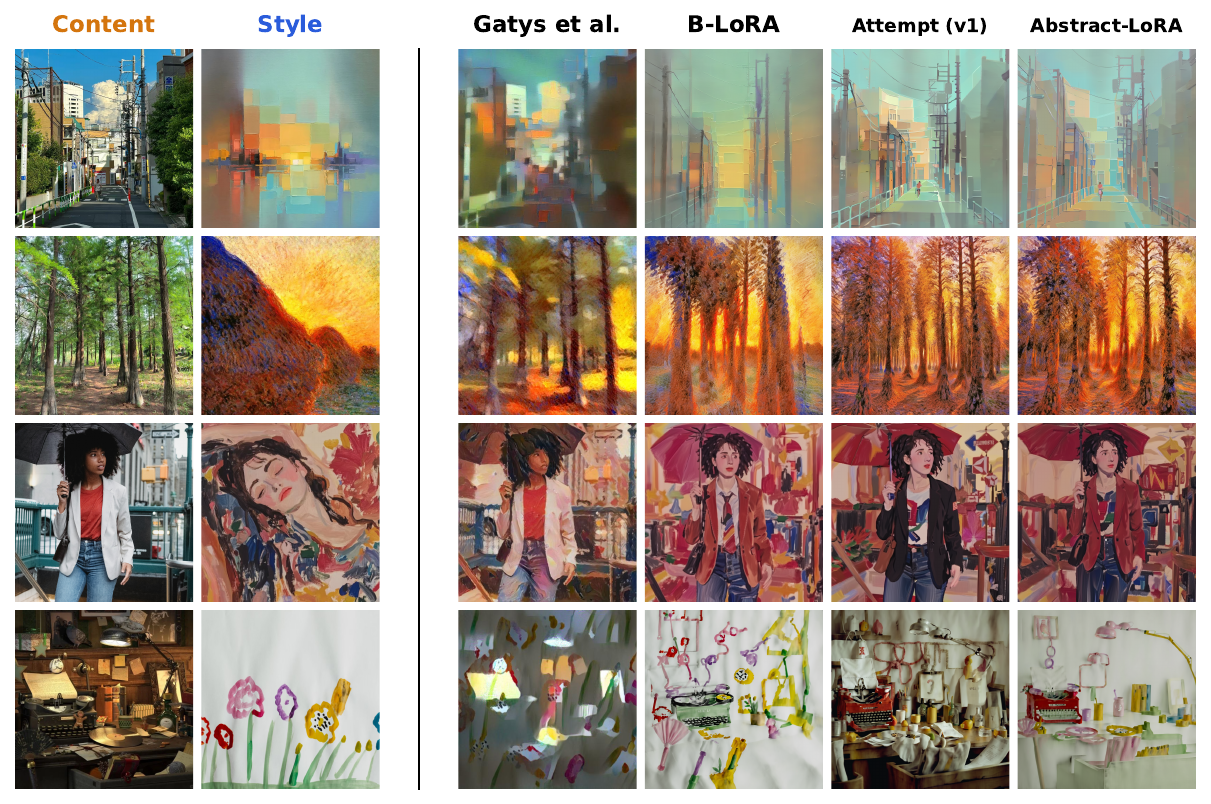}
    \caption{Rows 1-4 correspond to the results of a representative content-style pair from scenarios 1-4, respectively, across different methods.}
    \label{fig:results}
\end{figure}

\begin{table}[t]
\centering
\caption{Quantitative results of different methods in four scenarios}
\begin{tabular}{cc|cccc}
\toprule
\multicolumn{2}{c|}{} & Gatys et al. & B-LoRA & Attempt (v1) & Abstract-LoRA \\
\midrule
\multirow{3}{*}{scenario1} 
& VGG &0.901 &0.919 &0.872 &0.905 \\
& DINOv2 &0.311 &0.389 &0.543 &0.539 \\
& Total &1.212 &1.308 &1.415 &\textbf{1.444} \\
\midrule
\multirow{3}{*}{scenario2} 
& VGG &0.912 &0.904 &0.881 &0.902 \\
& DINOv2 &0.505 &0.408 &0.562 &0.553 \\
& Total &1.417 &1.312 &1.443 &\textbf{1.455} \\
\midrule
\multirow{3}{*}{scenario3} 
& VGG &0.724 &0.935 &0.861 &0.911 \\
& DINOv2 &0.617 &0.527 &0.571 &0.563 \\
& Total &1.341 &1.462 &1.432 &\textbf{1.474} \\
\midrule
\multirow{3}{*}{scenario4} 
& VGG &0.827 &0.933 &0.831 &0.912 \\
& DINOv2 &0.029 &0.491 &0.618 &0.573 \\
& Total &0.856 &1.430 &1.449 &\textbf{1.485} \\
\bottomrule
\end{tabular}
\label{tab:results}
\end{table}

\noindent In this section, we quantitatively evaluate the performance of our proposed method in terms of content preservation and style fidelity across various complex background scenarios.

\noindent\textbf{Evaluation Metrics}\quad (1) We employ VGG \cite{simonyan2015} to assess style-level preservation. VGG's hierarchical feature representations effectively capture low-level textures and high-level semantic patterns, making it well-suited for measuring stylistic similarity between generated and reference style images. (2) We utilize DINOv2 \cite{oquab2024} to evaluate content-level preservation. As a self-supervised vision transformer pre-trained on diverse image data, DINOv2 exhibits robust semantic understanding and is particularly effective at capturing object identity and structural information, making it ideal for assessing content fidelity.

\noindent\textbf{Baselines}\quad We use Gatys et al. \cite{gatys2016} and B-LoRA as our baselines, both implemented using their public code and default configurations. In addition, we report results for Attempt (v1) (Section~3.2), an intermediate version of our own method prior to incorporating the abstract representation, to isolate the contribution of this final component. As this is an early-stage exploration of the idea, our comparison currently centers on B-LoRA—the method our approach builds on most directly—rather than a broader survey of recent single-image style transfer techniques; we discuss this scope in more detail in Section~5.

Figure~\ref{fig:results} shows one representative content–style pair per scenario for qualitative comparison. The quantitative results in Table~\ref{tab:results}, however, are averaged over a broader evaluation set: 10 style images combined with 2 content images in each of the four scenarios, yielding 80 content–style pairs in total. Together, they illustrate the improvements our method offers over B-LoRA in both quantitative metrics and cross-scenario generalization.

The results reveal that incorporating the background LoRA matrix (Attempt (v1)) substantially improves content preservation over B-LoRA, as measured by DINOv2 scores, though at some cost to style fidelity, as measured by VGG scores. By further introducing abstract compensation, we recover most of this style fidelity—reaching levels comparable with B-LoRA—while incurring only a small additional reduction in content scores. As a result, Abstract-LoRA attains the highest total (VGG + DINOv2) score in all four scenarios among the methods we compare, suggesting a favorable overall balance between style fidelity and content preservation.

\section{Conclusion}

In this work, we present Abstract-LoRA, an approach that extends single-image style transfer through a more detailed decomposition of the U-Net architecture in diffusion models. By splitting content into background and subject components and identifying their corresponding U-Net blocks, our method addresses content preservation more thoroughly than B-LoRA, which treats content as a single, undifferentiated component. Introducing an abstract color representation further recovers style fidelity lost in this process, without resorting to naive weight scaling. Within the scope of our current experiments, Abstract-LoRA attains the highest total quantitative score among the compared methods in all four scenarios, and the ablation against Attempt (v1) suggests that the abstract representation meaningfully contributes to this improvement.

This work represents an early-stage exploration of the idea rather than a fully validated system. Our quantitative evaluation is currently limited to two baselines—Gatys et al. and B-LoRA. B-LoRA, however, is itself one of the more advanced single-image diffusion-based style transfer methods and was already benchmarked against a range of other approaches in its own evaluation, so comparing against it provides a meaningful, if indirect, point of reference for our method's standing in the broader landscape. Based on visual inspection, our results also compare favorably against other single-image style transfer methods reported in the literature, though we have not yet conducted a systematic qualitative or quantitative comparison against them. A more comprehensive evaluation against additional baselines, along with ablation studies over the method's hyperparameters, is needed to establish Abstract-LoRA's relative standing more rigorously, and will be the focus of future revisions of this work. We hope Abstract-LoRA nonetheless offers a useful starting point for thinking about background-aware disentanglement in single-image style transfer with diffusion models.

\section*{Code Availability}

The implementation and experimental code for Abstract-LoRA are publicly available at
\url{https://github.com/Xinglin-Hu/Abstract-LoRA}.

\bibliographystyle{unsrtnat}
\bibliography{references}

\appendix

\end{document}